# SDLNet: Statistical Deep Learning Network for Co-Occurring Object Detection and Identification

Binay Kumar Singh

Department of Computer Science, University of Central Florida, binay.singh@ucf.edu

Niels Da Vitoria Lobo

Department of Computer Science, University of Central Florida, niels.davitorialobo@ucf.edu

With the growing advances in deep learning based technologies the detection and identification of co-occurring objects is a challenging task which has many applications in areas such as, security and surveillance. In this paper, we propose a novel framework called SDLNet- Statistical analysis with Deep Learning Network that identifies co-occurring objects in conjunction with base objects in multilabel object categories. The pipeline of proposed work is implemented in two stages: in the first stage of SDLNet we deal with multilabel detectors for discovering labels, and in the second stage we perform co-occurrence matrix analysis. In co-occurrence matrix analysis, we learn co-occurrence statistics by setting base classes and frequently occurring classes, following this we build association rules and generate frequent patterns. The crucial part of SDLNet is recognizing base classes and making consideration for co-occurring classes. Finally, the generated co-occurrence matrix based on frequent patterns will show base classes and their corresponding co-occurring classes. SDLNet is evaluated on two publicly available datasets: Pascal VOC and MS-COCO. The experimental results on these benchmark datasets are reported in Sec 4.

## 1 Introduction

Object detection has shown tremendous growth in last decade due to its substantial popularity in various research areas of computer vision. The pioneer attempt for object detection based on deep learning initiated by the introduction of Region-based Convolutional Neural Network(RCNN) proposed by Girshick et al. in 2014 [1]. Following this a series of RCNN based models came into existence, such as Fast RCNN, Faster RCNN, Mask RCNN and so on. The primary role of object detector is to perform object detection apart from classification, segmentation, key point detection, scene understanding, and other tasks. To detect objects these ODs propose bounding boxes corresponding to each detected objects and predict labels for the bounding boxes.

In this paper we are extending the problem of multilabel multiclass detection and classification where we also want to find all the co-occurring object classes frequently occurring with base classes. So it is important to differentiate between base classes and co-occurring classes, to do so, in this problem setting, we used holistic approach to define base classes in terms of its frequency of appearance in the dataset, high frequency means more chance to include a class in base classes, contrary to this, for co-occurring objects we set a threshold value and if an object appears more than a threshold value accompanying base class then the object is considered as co-occurring object. The main intuition behind this is object when frequently occur with other objects chances are high that one object is related to another object or when objects share some common context, then they tend to appear more closely to each other. In other words, the reason behind co-occurrence objects frequently occur with base class could be due to its context sharing, natural placement, semantically related and so on.

Relating to this in humans, visual cortex especially para hippocampal cortex is responsible for representing strong contextual associations [3] with other objects or locations in co-occurring objects. For our intelligent agent, we can also use contextual knowledge generated from our surroundings that we people also use to identify objects, for example, looking at a computer desk our visual cortex captures all items on the desk like monitor, keyboard, mouse, and other stationary items. Another example might be base class is an apple and the co-occurring object might be a bowl, a knife or basket.

Our selection of state of the art object detection models are limited in the fact that here we need a model which can detect multiple objects in the images. Models which can perform only classification task [2] ruled out because these models typically perform well when single object is present in the image, while models based on detection and segmentation perform well on multilabel multiclass problems.

The motivation behind this research is the growing demand of computer vision based systems used in navigation and scene understanding. One interesting implementation of SDLNet is in security and surveillance. For example, consider a scenario at a crowded airport, one suspect is found now the security personnel want to know is there any other suspect related to this suspect, by using this approach it will be easy because the model will identify other person(co-occurrence class object) related to the main suspect(base class object). Details of which are provided in section 3. Some research towards learning co-occurrence statistics of object representation is done with the help of vision and language models [4], but in our research we are not implying any language models such as object2vec or word2vec [5]. The reason behind this is our proposed approach does not consider any form of contextual representation of objects.

In a nutshell, in this research work, we seek to develop a deep learning based model that can find all base class objects and their corresponding co-occurring objects and report the co-occurrence statistics in respect to base class objects. The proposed method can be described in a two stage pipeline as follows: the first stage of the pipeline deals with predicting class labels. To do so we used deep convolutional neural networks based feature extractor and followed by multiclass multilabel object detector and co-occurrence matrix that generates frequently occurring objects corresponding to each base classes.

In summary, our contributions in this paper are listed below.
- We proposed a new framework for co-occurring object detection and identification.
- We used feature extractors followed by classifier for multilabel multiclass classification and co-occurrence matrix analysis for frequently occurring objects in terms of base classes and co-occurring classes.
- Finally, we evaluate the proposed method on various evaluation metrics on two publicly available datasets.

## 2  Literature Review

The following subsections provide details on background information required for this research work.

### 2.1 Multilabel Object Detectors

Much research has been done in the field of object detection [6]. Multilabel object detection is a crucial task in computer vision aiming to identify and classify multiple objects within an image while allowing for the possibility of multiple objects belonging to different classes to be present simultaneously in one single image. This complex task finds applications in various domains such as autonomous driving, surveillance, medical imaging, and others. For any object detector the key components are feature extraction, object detection, classification, labelling, and post-processing. While developing an object detector there are some challenges researchers face, for example, label ambiguity, data imbalances, scalability, and real-time performance [7].

An object detector comes into either of these two categories: single stage object detectors or two stage object detectors, for example, Single Shot Detector(SSD), You Only Look Once(YOLO) are single stage detectors while RCNN, Fast RCNN, Faster RCNN, RetinaNet, and Detector Transformers(object detector based on transformer architecture) are in second category.

Faster RCNN introduced by S. Ren et al. in [8] significantly improved the speed and accuracy of object detection compared to previous approaches. Faster RCNN exploits the idea of RPN which generates bounding boxes that are likely to contain objects using a sliding window approach combined with anchor boxes. The RCNN takes these proposals as input and performs object classification and bounding box regression to define the proposals. The RPN shares convolutional layers with the subsequent RCNN, enabling end-to-end training. It predicts objectness scores and bounding box offsets for a set of anchor boxes at each spatial location. The anchor boxes are generated at different scales and aspect ratios to cover objects of various sizes and shapes.

Faster RCNN with MobileNetV3 is a lightweight model [9] much needed in resource-constrained environment, which makes it an ideal choice for deploying object detection models on mobile and embedded devices with limited computational resources such as our proposed framework. MobileNetV3 utilizes depthwise separable convolutions as its main building block by introducing Squeeze-and-Excitation(SE) blocks which factorizes standard convolutions into depthwise and pointwise convolutions. It introduces hard-swish activation function, which combines the advantages of the Swish activation function with the efficiency of the ReLU6 activation function.

Recent work based on transformers is introduced in DETR proposed in [10] performs better than two-stage detectors e.g. Faster RCNN- FPN+ and RetinaNet with 2x less FLOPS and APL is 61.1 an increase of 7.7 on large objects(subscript L represents large objects). DETR uses transformer's encoder and decoder network on top of extracted features from CNN and generates a fixed set(100) bounding boxes, and class labels. On the contrary for small objects its performance decreases -6.1AP from both two-stage detectors. This is due to the fact that attention mechanism use global reasoning capabilities in transformers.

EfficientDet [11] developed by Google backbone on EfficientNet model is a single stage detector perform better on edge devices. EfficientDet surpasses its peers on MS-COCO datasets. Meanwhile, all these object detectors are focused on tasks such as object detection, segmentation, etc. so these tasks are limited, in this paper we need a model that can do co-occurrence analysis as well.

Another line of work in object detection based on Transformer architecture is presented in [12]. Deformable DETR is designed to improve its performance, especially in handling small objects and achieving faster convergence during training. Deformable DETR integrates deformable convolutional layers into the standard DETR architecture to enhance its object detection capabilities. Deformable convolutions improve contextual modeling within the backbone network and transformer encoder, allowing Deformable DETR to capture long-range dependencies and relationships between objects more effectively.

## 2.2 Co-occurrence Object Prediction

Co-occurring object prediction involves identifying pairs or sets of objects that frequently appear together in a given context or dataset, and then predicting the occurrence of one object based on the presence of another. One of the challenges in multilabel multiclass problem is huge, annotated data that leads to problems such as learning about co-occurrence of objects, which we are dealing in this research paper. In [13] authors presents a semi-supervised based learning approach where for a trained network labels are used for inferring unlabeled images.

An interested idea presented in [14] where authors argue on Painting-Growth algorithms and N Painting-Growth algorithms which allows two items to scan at once to obtain the results. Experimental results suggest that it performs better than its predecessors.

Nowadays, in data mining also, researchers use frequency pattern analysis for identifying, [15]. In frequency pattern analysis a tree based approach is used formerly known as frequency pattern growth, [16] for mining data elements. It is worth mentioning that the frequency pattern growth is mostly used in data science. We followed a similar approach used in data science for co-occurring object prediction.

## 3 The Proposed Method

### 3.1 Overview of the Proposed Model

We have developed a method called SDLNet that can perform co-occurring object detection and prediction. We used deep leaning based object detectors for detecting objects and predicting labels and used statistical analysis based co-occurrence matrix for identifying base classes and corresponding co-occurring classes. Our proposed approach is presented in Figure 1 consists of multi-label object detection and classification and co-occurrence object prediction. The proposed approach is comprised of two stages discussed further.

The first stage consists of multilabel object detection and classification [17] and generating together base classes and frequently occurring classes. To do so, we used pre-trained object detector models Faster RCNN and DETR and fine-tune the last few layers based on the model to accommodate according to the datasets.

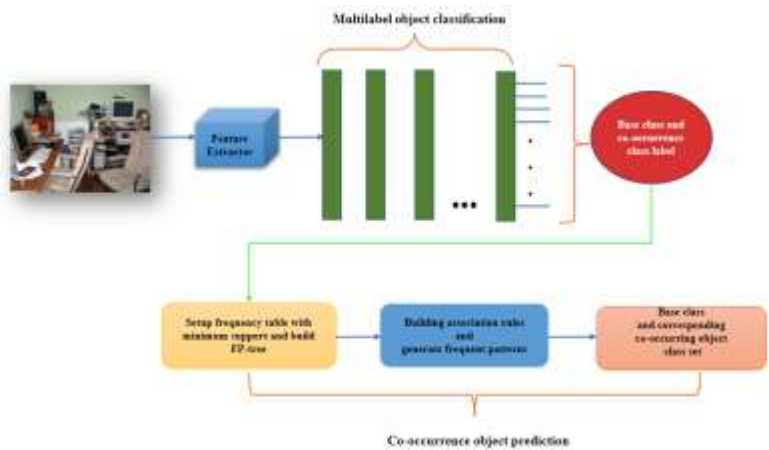

**Figure 1:** A unified feed-forward framework for co-occurrence object prediction. For each image in a dataset, the DCNN backbone feature extractor with multilabel object classifier generates a set of bounding boxes and labels that is further processed for base class and co-occurrence class generation.

The reason behind selecting these pre-trained models are performance and speed on detecting small and large size objects, and after carefully investigating some of the images in the datasets we found that frequently occurring objects are smaller in size compared to base class objects. To show the effectiveness of smaller and larger size models we used two different backbones of Faster RCNN networks- ResNet50 and MobileNetV3. The loss function is combining box prediction loss and class label logits. We performed various transformations on input images because of different sizes, details of which are provided in next section. After generating a set of class labels our next goal is to identify base classes and co-occurring classes.

In second stage we perform co-occurrence object analysis. In this stage, from the multilabel multiclass data we identify base classes and co-occurring classes(details are given in section 3.2). Base classes as defined earlier which must be present maximum number of times in the dataset, and co-occurring classes are those classes that appear frequently(greater than a minimum support value) corresponding to base classes.

## 3.2 Co-occurrence Label Prediction with Base Classes

As discussed earlier we perform co-occurrence label prediction in second stage of the pipeline. The co-occurrence object label prediction module [17] takes both base classes and frequently occurring classes at once and perform equation(1) operation by initializing frequently appearing label for base class.

$$L_1 = \{\{i\} \mid i \in I, \text{supp}(\{i\}) \geq \min\_\text{support}, \forall i\} \quad (1)$$

Here, $L_1$ represent the set of frequently labelsets of size 1, containing single item i with support greater than or equal to the minimum support threshold set 0.5. Following this we generate recursively co-occurring object candidate labelsets as shown in equation (2).

$$C_{k+1} = \begin{cases} \{i_1, i_2, \ldots, i_{k+1}\} \mid \{i_1, i_2, \ldots, i_k\} \in L_k, \\ \{i_1, i_2, \ldots, i_k\} \subset \{i_{k+1}\}, \\ \text{supp}(\{i_1, i_2, \ldots, i_{k+1}\}) \geq \min\_\text{support} \end{cases} \quad (2)$$

Here, $C_{(k+1)}$ is the set of candidate labelsets of size $k+1$, that filters out labelsets that do not meet the minimum threshold. Because there are many candidate labelsets generated that may not be part of final co-occurring object sets, so we prune infrequent candidate labelsets as per equation(3).

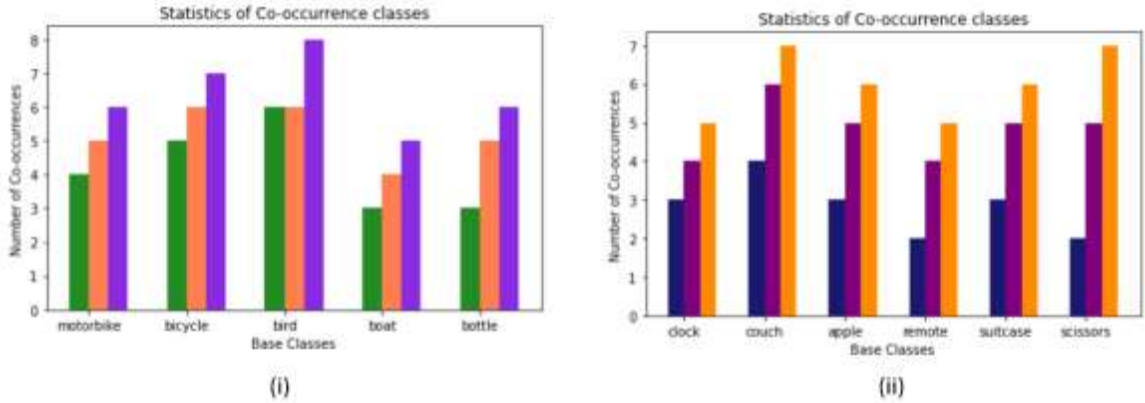

**Figure 2:** This chart represents class specific co-occurrences on datasets (i) MS-COCO (ii) Pascal VOC. These are evaluated on all the three models discussed in section 4. Here, x-axis represents base classes and y-axis represents number of co-occurrences corresponding to base classes.

$$L_{k+1} = \begin{cases} \{i_1, i_2, \ldots, i_{k+1}\} \mid \{i_1, i_2, \ldots, i_k\} \in C_{k+1}, \\ \text{supp}(\{i_1, i_2, \ldots, i_{k+1}\}) \geq \min\_\text{support} \end{cases} \quad (3)$$

## 4 Experiments and results

For co-occurring object detection and prediction we need dataset that must contain multiple labels in images so have chosen two object detection datasets Pascal VOC-2012 and MS-COCO-2017. These datasets are widely used for image classification and segmentation tasks. Pascal VOC contains 20 classes and have 11,540 images in 'trainval' set and 10,991 images in 'test' set, while MS-COCO contains 80 categories and contains 118,287 images in 'train' set and 40,670 images in 'test' set, each dataset contains multiple bounding boxes and have multiple objects from multiple scenes.

## 4.1 Experimental Settings

We implemented SDLNet in Pytorch and used Intel i7 processor, NVIDIA GeForce RTX 2060 GPU with 16GB RAM. We used optimizer as 'Adam' and run the experiment for 30 epochs and considered learning rate 0.01. Since both datasets contain different resolution images, for example Pascal VOC-2012 dataset have a resolution of 1280 x 720 or 1600 x 1200, so we decided to resize all images to 224 x 224. We performed various transformations such as random horizontal flip, random rotation and normalized the dataset on ImageNet1k mean and standard deviation for better performance.

In stage 1 of the pipeline we used a pretrained model Faster RCNN with two backbones ResNet50 and MobileNetV3 and retrained last few layers only. Another popular model based on transformers used here is DETR. We used the same settings discussed earlier for both models.

In figure 2, we have shown performance of these models on two datasets. We empirically considered these classes(shown in x-axis) from the datasets and plotted their frequently occurring classes. First tick represents Faster RCNN on backbone ResNet50, second tick represents Faster RCNN on backbone MobileNetV3 and third tick represents DETR in both charts. It can be observed from both charts that DETR identifies more objects than others in both datasets. Our figure 3 shows number of co-occurrence classes irrespective of base classes. As from this figure we can understand that the maximum number of labels identified for a base class is 8.

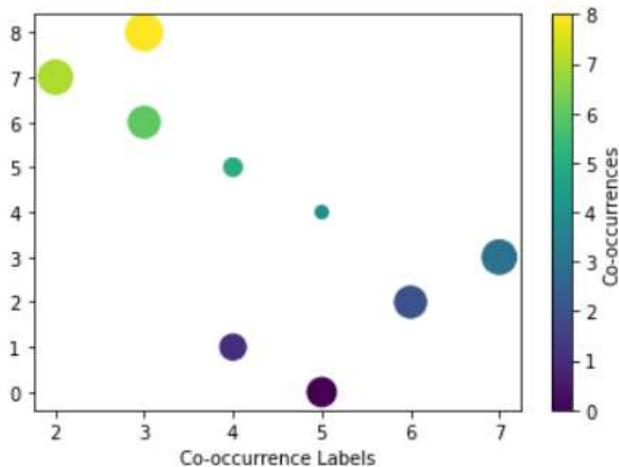

**Figure 3:** Number of co-occurrences of classes, irrespective of base classes.

## 4.2 Results

In terms of reporting performance of the first part of pipeline we used two different metrics average precision and mAP(mean Average Precision), mAP is calculated based on the below equation(4).

$$\mathrm{mAP} = \frac{1}{N}\sum_{i=1}^{N}\mathrm{AP}_i \qquad (4)$$

Here, Table 1 shows performance of three different models/backbone. From this it is clear that DETR performed better than other models. For both dataset DETR mAP is 68.8 and 68.9.

Furthermore, after reporting the model performance in Table 1, we discard 'reg' layer output and consider only 'cls_score' head. This 'cls_score' is used further for identifying base classes and their corresponding co-occurring classes.

**Table 1: Model performance on various parameters**

| Datasets | Models/Backbone | AP | mAP |
|---|---|---|---|
| Pascal VOC(2012) | ResNet50 | 61.5 | 63.3 |
|  | MobileNetv3 | 62.2 | 64.1 |
|  | DETR | 65.3 | 68.8 |
| MS-COCO(2017) | ResNet50 | 62.7 | 64.2 |
|  | MobileNetV3 | 63.9 | 65.3 |
|  | DETR | 66.4 | 68.5 |

Since the object detectors predict a set of class labels and the later part of the framework for predicting base classes and frequently occurring classes are solely dependent on the earlier part, thereby, we concluded that performance of our proposed framework relies on the performance of object detectors as well. We also observed that the performance of the entire framework is centric towards various ODs and statistical analysis.

## 5  Conclusion

We have presented a new framework for identifying frequently occurring objects with respect to base classes. We proposed the method in two stages where in the first stage we locate all the objects present in the image and their corresponding labels, and in the second stage we find all base classes and their corresponding co-occurring classes set by a threshold value. For first stage we used Faster RCNN based architecture and Transformers based architecture and in the second stage we performed statistical analysis based on pattern analysis. We used datasets Pascal -VOC and MS-COCO containing multiple objects in multiple images in our experiments and reported results. Further, we plan to extend this work and consider co-occurring classes as unknown or occluded and discover these unknown or occluded classes along with base classes.